\documentclass[letterpaper, 10 pt, conference]{ieeeconf}
\IEEEoverridecommandlockouts                              % This command is only needed if 
\usepackage{graphics} % for pdf, bitmapped graphics files
\usepackage{epsfig} % for postscript graphics files
\usepackage{mathptmx} % assumes new font selection scheme installed
\usepackage{times} % assumes new font selection scheme installed
\usepackage{amsmath} % assumes amsmath package installed
\usepackage{amssymb}  % assumes amsmath package installed
\usepackage{color}
\usepackage{subfig}
\usepackage[dvipsnames]{xcolor}

\usepackage{soul} % Please remove before submission
\newcommand{\cosX}{cX}
\newcommand{\cosY}{cY}
\newcommand{\cosZ}{cZ}
\newcommand{\cosXdot}{c\dot{X}}
\newcommand{\cosYdot}{c\dot{Y}}
\newcommand{\cosZdot}{c\dot{Z}}

\title{\LARGE \bf
Real-time Uncertainty-Aware Motion Planning for Magnetic-based Navigation
}

\author{Aditya Penumarti$^{1}$, Kristy Waters$^{1}$, Humberto Ramos$^{1}$, Kevin Brink$^{2}$ and Jane Shin$^{1}$% <-this % stops a space
\thanks{*This work was supported by Air Force Research Lab Contract FA8651-22-F-1052 and Grant FA8651-23-1-0003.}% <-this % stops a space
\thanks{$^{1}$Aditya Penumarti, Kristy Waters, Humberto Ramos, and Jane Shin are with Department of Mechanical and Aerospace Engineering,
        University of Florida, Gainseville, FL 32611, USA
        {\tt\footnotesize \{apenumarti, watersk, jramoszuniga, jane.shin\}@ufl.edu }}%
\thanks{$^{2}$Kevin Brink is with the Air Force Research Lab,
        Eglin AFB, FL 32542, USA
        {\tt\footnotesize kevin.brink@us.af.mil}}%
}

\begin{document}

\maketitle
\thispagestyle{empty}
\pagestyle{empty}

%%%%%%%%%%%%%%%%%%%%%%%%%%%%%%%%%%%%%%%%%%%%%%%%%%%%%%%%%%%%%%%%%%%%%%%%%%%%%%%%
\begin{abstract}
Localization in GPS-denied environments is critical for autonomous systems, and traditional methods like SLAM have limitations in generalizability across diverse environments. Magnetic-based navigation (MagNav) offers a robust solution by leveraging the ubiquity and unique anomalies of external magnetic fields. This paper proposes a real-time uncertainty-aware motion planning algorithm for MagNav, using onboard magnetometers and information-driven methodologies to adjust trajectories based on real-time localization confidence. This approach balances the trade-off between finding the shortest or most energy-efficient routes and reducing localization uncertainty, enhancing navigational accuracy and reliability. The novel algorithm integrates an uncertainty-driven framework with magnetic-based localization, creating a real-time adaptive system capable of minimizing localization errors in complex environments. Extensive simulations and real-world experiments validate the method, demonstrating significant reductions in localization uncertainty and the feasibility of real-time implementation. The paper also details the mathematical modeling of uncertainty, the algorithmic foundation of the planning approach, and the practical implications of using magnetic fields for localization. Future work includes incorporating a global path planner to address the local nature of the current guidance law, further enhancing the method’s suitability for long-duration operations.
\end{abstract}

%%%%%%%%%%%%%%%%%%%%%%%%%%%%%%%%%%%%%%%%%%%%%%%%%%%%%%%%%%%%%%%%%%%%%%%%%%%%%%%%
\section{INTRODUCTION}

% (1) gps denied navigation is important, but under dynamic environment is very difficult because of the current sensing capabilities.
Localization is a critical component of autonomous systems, providing the necessary spatial awareness for navigation and task execution. In GPS-denied environments, where traditional satellite-based localization is unreliable or unavailable, ensuring precise localization becomes particularly challenging. Simultaneous localization and mapping (SLAM) or visual-SLAM methods have been extensively studied in these contexts. These methods use LiDAR measurements or computer vision algorithms to detect and utilize environmental features for localization. Other approaches employ sensing mechanisms such as Wi-Fi, Bluetooth, or inertial measurements \cite{huang_indoor_2023}. While these methods are highly reliable in certain environments, they are not generalizable as they are coupled to specific environments and applications \cite{pascacio_collaborative_2021}. For example, it is difficult to generalize these methods in environments that lack visual features or have dynamic features, where sensors cannot function as expected due to physical properties (e.g., underwater), or when missions need to be performed on a larger scale or for long durations \cite{ouyang_survey_2022}.

\begin{figure}
    \centering
    \includegraphics[width=\linewidth]{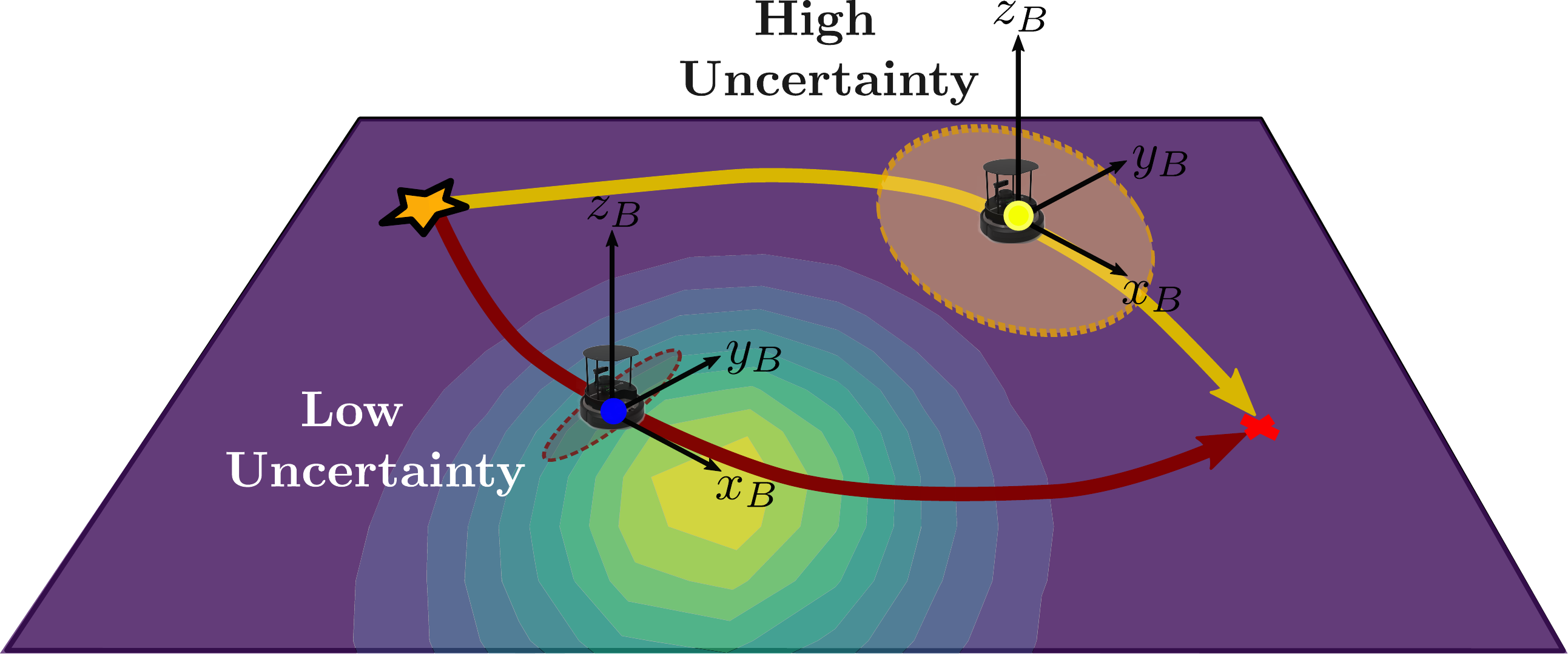}
    \caption{A representative scenario of uncertainty-aware guidance for magnetic-based navigation
    %The robot navigating through the area of low information has less certainty of its current state than the navigation in the high information gradients that increase the certainty of the state. This uncertainty of state is represented by the ellipses drawn around the robots, representing the state covariance\js{**}
    }
    % revisit the figure and try other ideas
    \label{fig:reference}
    \vspace*{-0.5cm}
\end{figure}

% (2) Magnetic navigation is one of the solution that can tackle that limitaciotns. these many literature has shown that magnetic navigation is the potential next important sensing modality for gps denied navigation.
Magnetic-based navigation (MagNav) is a promising solution to tackle these challenges \cite{gnadt2020signal}. Leveraging the robustness and uniqueness of external magnetic fields, MagNav has gained popularity due to the availability of magnetic anomaly maps and the resilience of magnetometers against jamming compared to radio-based devices \cite{canciani_airborne_2017}. MagNav utilizes field gradients or high spatial frequency features for localization \cite{canciani_absolute_2016} and has successfully been implemented on both ground \cite{storms_magnetic_2010, shockley_ground_2012, ramos_information-aware_2022} and aerial \cite{canciani_absolute_2016, canciani_airborne_2017, canciani_magnetic_2022} vehicles in indoor and outdoor environments. Successful MagNav requires basic navigational priors, such as a magnetic anomaly map, a motion model, and magnetic intensity measurements. While localization without a prior map is possible via SLAM, the results tend to outperform with prior maps. Researchers have developed various approaches, such as Gaussian-Process regression \cite{vallivaara_simultaneous_2010} and informative routing \cite{kemppainen_magnetic_2015}, to enhance MagNav's accuracy using SLAM, which has been applied to an airborne data set \cite{lee_magslam_2020}. However, localization errors in the results suggest that additional sensors might be necessary for effective localization.

% (3) While it is promising, however, there has not been much of work on utilizing this map into the planning to reduce the localization uncertainty.
Given a magnetic field map, MagNav can utilize Bayesian estimation techniques, like particle or Kalman filters, for localization \cite{canciani_absolute_2016}. However, instead of just using the map for localization, limited work has been done on harnessing the unique information from the map into active planning algorithms in order to improve localization. By integrating the map information into planning, information-driven methods, such as active navigation, can consider localization uncertainty directly and guide the vehicle to the path that can minimize the uncertainty. There exist previous research that considers localization uncertainty in MagNav. For example, in \cite{quintas_auv_2019}, Cramer Rao Lower Bound (CRLB) is used to quantify path uncertainty; however, the uncertainty is not directly taken into account in planning. In \cite{ramos_information-aware_2022}, non-linear observability-based and information-driven navigation methodologies are proposed; however, the observability approach struggles with its insensitivity to uncertainty parameters \cite{rafieisakhaei_use_2017}, and the computational complexity of the expected entropy reduction (EER) calculation hinders hardware testing \cite{burgard_active_1997-1, shin_imvp_22}. These challenges highlight the need for efficient uncertainty-aware methods in MagNav to optimize navigation accuracy and efficiency in GPS-denied environments.

% (4) So in this paper, we propose an uncertainty-aware motion planning algorithm that can be used for magnetic navigation. Our work does ... (why the presented work is impactful)
Therefore, this paper proposes a real-time uncertainty-aware motion planning algorithm for magnetic navigation. Magnetic fields, with their ubiquity and distinct local anomalies, provide reliable localization cues. By equipping robots with onboard magnetometers, we develop an information-driven planning methodology that plans trajectories based on real-time localization confidence. This approach considers the trade-off between finding the shortest or most energy-efficient routes and reducing localization uncertainty for enhancing navigational accuracy and reliability. The contributions include integrating an uncertainty-driven navigation framework with magnetic-based localization, resulting in a real-time adaptive system capable of minimizing localization errors in complex environments. Additionally, the presented method is validated through extensive simulations and real-world experiments, demonstrating its performance in reducing localization uncertainty and the feasibility of real-time implementation. This paper details the mathematical modeling of uncertainty, the algorithmic foundation of the presented path planning approach, and the practical implications of using magnetic fields for localization, highlighting the potential of uncertainty-aware navigation to revolutionize autonomous systems.

\section{PROBLEM FORMULATION}
This paper considers a motion planning problem for a mobile robot tasked with navigating in a GPS-denied environment while minimizing its own localization uncertainty using onboard one vector and one scalar magnetometer. The workspace is defined by a 2-dimensional space $\mathcal{W}\subset \mathbb{R}^{2}$, and the magnetic field map over $\mathcal{W}$ is denoted by $m~:~\mathcal{W}\rightarrow \mathbb{R}^{+}$. The magnetic field map $m$ is assumed known \textit{a priori}, for example, from pre-surveying, and static. Localization is performed solely based on the onboard magnetometer measurement and a given magnetic field map, without other sensors like internal measurement unit.

\begin{figure}
    \centering
    \includegraphics[width=0.95\linewidth]{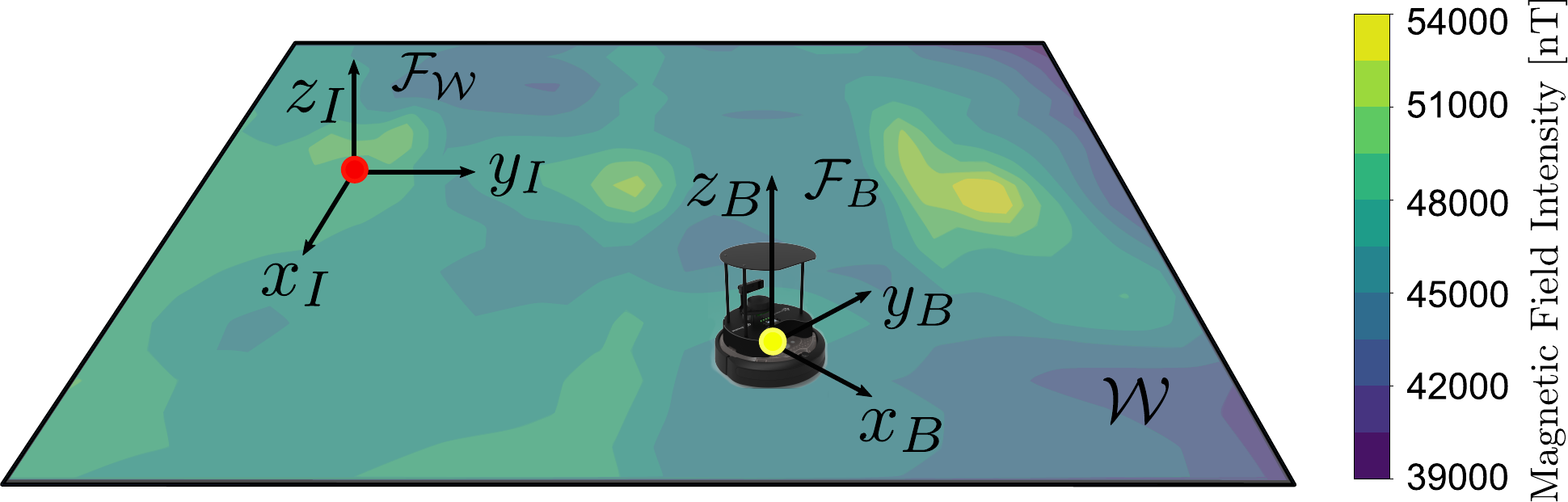}
    \caption{Schematic of the workspace and coordinate systems}
    % increase the font size and reduce the color bar height to match the workspace figure (I think it will save time and put III to the beginning of the next page)
    \label{fig:schematic}
\end{figure}

% robot state
As illustrated in Fig. \ref{fig:schematic}, the inertial frame $\mathcal{F}_{W}$ with $x_{I}y_{I}z_{I}$-axes is defined in $\mathcal{W}$, and the body-fixed frame $\mathcal{F}_{B}$ with $x_{B}y_{B}z_{B}$-axes is defined and fixed on the mobile robot. The robot position and orientation is defined by the position and orientation of $\mathcal{F}_{B}$ with respect to $\mathcal{F}_{W}$. In this paper, we assume that the mobile robot is a differential-drive ground robot, although this assumption can be easily generalized by considering other dynamics constraints. Then, the position of the robot is defined by $x$ and $y$ position of $\mathcal{F}_{B}$ in $x_{I}y_{I}$-plane, and the orientation of the robot is defined by $\theta$, which is the angle of $x_{B}$-axis with respect to $x_{I}$-axis. By denoting a certain time step by $k=0,1,2,\dots$, a state vector that represents the robot pose at time $k$ is defined by $\mathbf{x}_{k}=[x_{k}~y_{k}~\theta_{k}]^{T}$.

% robot motion model
By denoting the control state vector at time step $k$ as $\mathbf{u}_{k}$, the robot's motion model is denoted by
\begin{equation}
    \mathbf{x}_{k+1} = \mathbf{f}\left(\mathbf{x}_{k}, \mathbf{u}_{k}\right)
    \label{eq:motion_model}
\end{equation}
In this letter, the motion model of a differential-drive robot can be derived from a unicycle model. By letting $V$ and $\omega$ be the linear and angular velocities, the kinematics of a unicycle model gives
\begin{align}
    x_{k+1} &= x_{k} +V\cos \theta \Delta t + \epsilon_{x}\\
    y_{k+1} &= y_{k} + V \sin \theta \Delta t + \epsilon_{y}\\
    \theta_{k+1} &= \theta_{k} + \omega \Delta t + \epsilon_{\theta}
\end{align}
where $\Delta t$ is the size of each time step and $\epsilon_{x}$, $\epsilon_{y}$, $\epsilon_{\theta}$ are random Gaussian noises with zero mean and standard deviations $\sigma_{x}$, $\sigma_{y}$, and $\sigma_{\theta}$, respectively. The robot's control input is defined by $\mathbf{u}_k = \left[V~\omega \right]^{T}$. This motion model can be represented by a probability distribution $p(\mathbf{x}_{k+1}|\mathbf{x}_{k},\mathbf{u}_{k})$.

% sensor model - need to reformulate including basics
The sensor measurement model represents the process by which magnetometer measurements are being generated. The magnetometer reading obtained from the state $\mathbf{x}_{k}$ is denoted by $z_{k}\in\mathbb{R}^{+}$. By assuming that this reading is only dependent on the earth magnetic field and robot's current pose, the sensor model can be represented by
\begin{equation}
    z_{k} = g(\mathbf{x}_{k},m)
    \label{eq:sensor_model}
\end{equation} 
The measurement model can be defined by a probability distribution conditioned on the target state $\mathbf{x}_{k}$ and the map $m$ and denoted by $p(z_{k}|\mathbf{x}_{k},m)$. As an example, one can model the sensor measurement model as a Gaussian by $z_{k} \sim \mathcal{N}\left(m(\mathbf{x}_{k}),\sigma^2\right)$, which represents a magnetometer that obtains the magnetic field intensity at $\mathbf{x}_{k}$ with a Gaussian noise with variance $\sigma^2$.

% so this is the problem
The motion planning problem of a mobile robot navigating in a GPS-denied environment can be formulated as finding a sequence of control inputs $\mathbf{u}_{k}$ such that the uncertainty associated with the robot localization is minimized while the robot is navigating from a given initial position to a given goal position. The robot localization can be performed by updating the probability distribution, or belief, $p(\mathbf{x}_{k+1}|\mathbf{x}_{0:k},\mathbf{u}_{0:k},z_{0:k},m)$, where the subscript $(\cdot)_{0:k}$ denotes a history of the variable from time $0$ to $k$. The motion planning algorithm must compute the control inputs that can minimize the uncertainty associated with this probability distribution by considering the motion model, sensor measurement model, and the magnetic field map given \textit{a priori}.

\section{UNCERTAINTY-AWARE MAGNETIC-BASED NAVIGATION}
\label{sec:method}
This section describes the novel robot motion planning methodology for GPS-denied navigation using onboard magnetometers and a pre-collected magnetic field map. The presented method consists of four main components: (\textit{i}) Tolles-Lawson calibration, (\textit{ii}) Bayesian filter-based localization, (\textit{iii}) uncertainty-driven motion planning algorithm, and (\textit{iv}) particle-based approximation of expected entropy reduction for real-time implementation.

\subsection{Tolles-Lawson Magnetic Calibration}
\label{ssec:tolles-lawson}
% tolles-lawson model
The foundational step in the presented methodology involves the Tolles-Lawson (TL) calibration, which isolates and corrects the distortions in the magnetometer readings influenced by the robot's own motion and its onboard electronics. This calibration is critical for localization and motion planning because the robot makes decisions using the information on the external magnetic environment, not affected by its internal factors. The conventional TL model assumes the platform housing the magnetometer influences Earth's magnetic field readings ($B_{e}$) due to three main constant sources: permanent magnetization of the platform ($B_{p}$), induced fields due to magnetically susceptible material interacting with the Earth's field ($B_{i}$), and eddy current magnetic fields ($B_{ed}$) in conductive materials. For indoor navigation, the augmented TL \cite{ramos_ownship_nodate} is applied to account for strong nearby biases ($B_{bias}$) and the linear velocity effects ($B_{vel}$). Then, the total magnetic field reading, reported by the magnetometer, is represented by
\begin{equation}
    B_{t} = B_{e} + B_{p} + B_{i} + B_{ed} + B_{bias} + B_{vel}
    \label{eq:Bt_augmented}
\end{equation}

The calibration step aims to obtain $B_{e}$ from the sensor measurement $B_{t}$ by finding and eliminating the other terms. The scalar magnetometer gives $B_{t}$ value, and the vector magnetometer gives each component $B_x$, $B_y$, and $B_z$. These values are used to find the direction cosines terms, $\cos X$, $\cos Y$, and $\cos Z$, by
\begin{equation}
    \begin{bmatrix}
        \cfrac{B_x}{B_{t}}~ \cfrac{B_y}{B_{t}}~ \cfrac{B_z}{B_{t}}
    \end{bmatrix}^T
    = \begin{bmatrix}
        \cos X~ \cos Y~ \cos Z
    \end{bmatrix}^T
    \label{eq:dir_cosines}
\end{equation}
By denoting $\cos(X)$as $cX$ and applying the same rule for simplicity, we can find the rest of terms from TL as
\begin{align}
    \begin{split}
        & B_{p}= \epsilon_1 \cosX+\epsilon_2 \cosY+\epsilon_3 \cosZ\\
        & B_{i} = B_{t} ( \epsilon_4 \cosX^2 + \epsilon_5 \cosY^2 + \epsilon_6 \cosZ^2 + \epsilon_7 \cosX \cosY + \epsilon_8 \cosX \cosZ + \epsilon_9 \cosY \cosZ )\\
        & B_{ed} = B_{t}\left(\epsilon_{10} \cosX \cosXdot + \epsilon_{11} \cosX \cosYdot +\epsilon_{12} \cosX \cosZdot +\epsilon_{13} \cosY \cosXdot +\epsilon_{14} \cosY  \right. \\
        & \quad \quad \quad \left. \epsilon_{15} \cosY \cosZdot \cosYdot +\epsilon_{16} \cosZ \cosXdot + \epsilon_{17} \cosZ \cosYdot   +\epsilon_{18} \cosZ \cosXdot \right)\\
        & B_{bias} = \epsilon_{19} \text{ and } B_{vel}=\epsilon_{20}V
    \end{split}
    \label{eq:aug_TL}
\end{align}
% \begin{multline}
%     B_{p}=  \epsilon_1 \cosX+\epsilon_2 \cosY+\epsilon_3 \cosZ\\
%     B_{i} =  B_{t} ( \epsilon_4 \cosX^2 + \epsilon_5 \cosY^2 + \epsilon_6 \cosZ^2 + \epsilon_7 \cosX \cosY + \epsilon_8 \cosX \cosZ + \epsilon_9 \cosY \cosZ )\\
%     B_{ed}=  B_{t}\left(\epsilon_{10} \cosX \cosXdot + \epsilon_{11} \cosX \cosYdot + \epsilon_{12} \cosX \cosZdot  +\epsilon_{13} \cosY \cosXdot +\epsilon_{14} \cosY \right. \\
%         \left. +\epsilon_{15} \cosY \cosZdot \cosYdot +\epsilon_{16} \cosZ \cosXdot + \epsilon_{17} \cosZ \cosYdot + \epsilon_{18} \cosZ \cosXdot \right)\\
%     B_{bias}=\epsilon_{19} \text{ and } B_{vel}=\epsilon_{20}V
%     \label{eq:aug_TL}
% \end{multline}
where $\epsilon_i$, for $i=1,\dots 20$, are the calibration parameters that can be computed using least square method.

\begin{figure}
    \centering
    \includegraphics[width=0.9\linewidth]{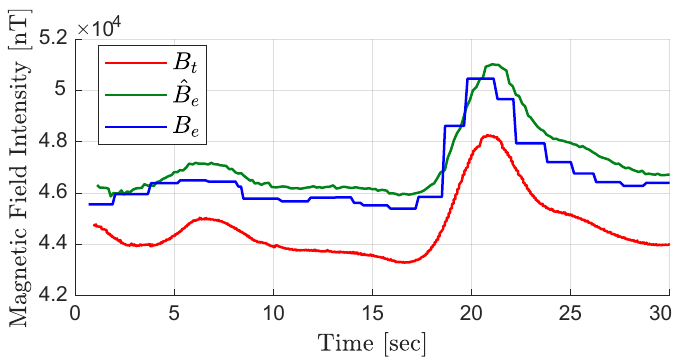}
      \caption{The Tolles-Lawson calibration result ($\hat{B}_{e}$) compared with the ground truth ($B_{e}$) and raw sensor readings ($B_{t}$)
      % Tolles-Lawson calibrated versus uncalibrated total field magnetic values. The reference field value is not continuous and stepped as the reference map collected in a grid format without interpolation.
      }
      % explain about the notation. change Btc into hat B_t from TL
    \label{fig:tolles_lawson_cal_vs_uncal}
\end{figure}

Denoting all the calibration parameters by a vector $\mathbf{\epsilon}=[\epsilon_1,\epsilon_2\hdots,\epsilon_{20}]^T$, we can re-write \eqref{eq:Bt_augmented} by $B_{t}-B_{e}=A\mathbf{\epsilon}$, where $A$ is a row vector consisting of the directional cosine terms from measurements and the constant linear velocity $V$ from \eqref{eq:aug_TL}. By vertically concatenating this equation for $m$ collected measurements, the calibration step is represented by $\mathbf{B}_{t}-\mathbf{B}_{e} = \mathbf{A}\cdot\mathbf{\epsilon}$, where each row of $\mathbf{A}$ contains one magnetic reading at a certain pose. Then, the least squares solution for $\mathbf{\epsilon}$ becomes
\begin{equation}
    \boldsymbol{\epsilon}=(\mathbf{A}^{\mathrm{T}}\mathbf{A})^{-1}\mathbf{A}^{\mathrm{T}}(\mathbf{B_{t}}-\mathbf{B_{e}})
    \label{eq:least_square}
\end{equation}
The calibration process and result are illustrated in Figure \ref{fig:tolles_lawson_cal_vs_uncal}.

% JS: here are the things to note. Where should we include these?
% It is important to note that we use $B_{t}$ from the scalar magnetometer, not the 2-norm of $B_x$, $B_y$, and $B_z$ from vector magnetometer. This approach is preferred due to the superior accuracy of modern scalar magnetometer magnitude readings compared to vector magnetometer readings. 
% Note that, as recommended in this augmented TL approach, 1) we apply a Butterworth filter to the magnetometer signals (with a cutoff frequency of 10 Hz) and 2) retain only the direction information from the vector magnetometer.
% Note that $B_{e}$ is the ``clean'' reference magnetometer reading, ideally the Earth's magnetic field without disturbances. To collect the reference values $B_{e}$ in practice, one requires a vehicle with the magnetometer mounted far from any non-earth magnetic effects. However, the magnetic readings become easily contaminated for indoor (our case) MagNav calibration on small platforms. For this reason, even when the magnetometer placement may be optimized to minimize disturbances, $B_{e}$ may not perfectly report the Earth's magnetic field. Nonetheless, this paper considers the cleanest $B_{e}$ that can be collected as the reference magnetic field $B_{e}$. For this paper's experiments, the reported $B_{e}$ corresponds to measurements taken when the robot is at rest and only carrying bare-minimum electronics.

\subsection{Localization in Magnetic-based Navigation}
The robot localization is performed using Bayes filter based on the probabilistic motion and sensor models defined in \eqref{eq:motion_model} and \eqref{eq:sensor_model}. Specifically, the posterior distribution $p(\mathbf{x}_{t}|z_{1:t}, \mathbf{u}_{1:t})$ is calculated from the corresponding posterior at the previous time step $p(\mathbf{x}_{t-1}|z_{1:t-1}, \mathbf{u}_{1:t-1})$ \cite{thrun_probabilistic_2005}. Following the notation in \cite{thrun_probabilistic_2005}, we denote this posterior by
\begin{equation}
    bel(\mathbf{x}_{k}) = p(\mathbf{x}_{t}|z_{1:k}, \mathbf{u}_{1:k}),
    \label{eq:belief}
\end{equation}
where $bel(\mathbf{x}_{k})$ is the belief of the state and will be used interchangeably with $bel(\mathbf{x}_{k}\mid z_k)$ for clarification purposes to indicate that a sensor measurement is used. From Bayes rule, Markov assumption, and state completeness, this posterior is updated recursively at every time step after executing control input $\mathbf{u}_{k-1}$ and obtaining the measurement $z_{k-1}$ as
\begin{equation}
    bel(\mathbf{x}_{k}\mid z_k) = \eta p(z_{k}|\mathbf{x}_{k},m) \int p(\mathbf{x}_{k}|\mathbf{x}_{k-1}, \mathbf{u}_{t}) bel(\mathbf{x}_{k-1})d\mathbf{x}_{k-1}
    \label{eq:bayesfilter}
\end{equation}
where $\eta$ is a normalizing constant.
% should I justify the assumptions here?
It is notable that, in the presented magnetic-based navigation setting, the measurement model used in the filter includes the map information as shown in \eqref{eq:bayesfilter}.

\subsection{Uncertainty-Aware Motion Planning}
\label{ssec:eer_planning}
For the planning algorithm to consider localization uncertainty, the presented method defines the uncertainty as a differential entropy to consider a continuous state. The differential entropy is defined based on the posterior at the current time step, $bel(\mathbf{x}_{k})$, by
\begin{equation}
    H\left(bel\left(\mathbf{x}_{k}\right)\right) = -\int_{\mathbf{x}_{k}} bel\left(\mathbf{x}_{k}\right) \cdot \log \left(bel\left(\mathbf{x}_{k}\right)\right) d \mathbf{x}_{k}
    \label{eq:entropy}
\end{equation}
where $\log$ is denoting $\log_{2}$ for simplicity. Then, the information gain, i.e., entropy reduction, is defined by
\begin{equation}
    I\left(z_{k}, \mathbf{u}_{k} \right) = H\left( bel\left(\mathbf{x}_{k-1} \right) \right) - H\left(bel\left(\mathbf{x}_{k} \mid \mathbf{u}_{k}, z_{k} \right) \right)
    \label{eqn:information_gain}
\end{equation}
Note that the last term $bel\left(\mathbf{x}_{k} \mid \mathbf{u}_{k}, z_{k} \right)$ is identical to $bel(\mathbf{x}_{k})$, which is computed via Bayesian update in \eqref{eq:bayesfilter}, and the conditioned variables are explicitly written for clarification purposes. Then, the reduction of differential entropy is used as an information gain function.

% EER definition
To consider the reduction of localization uncertainty, the presented motion planning algorithm is designed to find the control input that gives the maximum reduction in this differential entropy. Since entropy reduction can only be calculated after updating the posterior after obtaining a measurement, the planning algorithm should compute the expected entropy reduction (EER) value as a function of control input. Therefore, EER is defined to predict future uncertainty reduction by
\begin{multline}
    EER\left( \mathbf{u}_{k+1}\right) = \mathbb{E}_{z_{k+1}} \left[ I\left(z_{k+1}, \mathbf{u}_{k+1} \right) \right]\\
    = \int_{z_{k+1}} p\left(z_{k+1} \mid \mathbf{x}_{k}, \mathbf{u}_{k+1} \right) I\left(z_{k+1}, \mathbf{u}_{k+1}\right) dz_{k+1}
    \label{eqn:EER}
\end{multline}
where $p\left(z_{k+1}\mid \mathbf{x}_{k}, \mathbf{u}_{k+1}\right)$ is the probability distribution of the future sensor measurement at the future robot pose $\mathbf{x}_{k+1}$ after executing the control input $\mathbf{u}_{k+1}$. %This EER value is basically an average of future entropy reduction values computed by considering all possible future sensor measurements.

% cost function
Then, the final cost function is designed to consider the trade-off between minimizing localization uncertainty and reducing the total travel time. First, the action space is defined by a set of $m$ possible discretized control inputs $A = \left\{ \mathbf{a}_{1}, \mathbf{a}_{2}, \dots, \mathbf{a}_{m} \right\}$ where $\mathbf{a}_{j}$, for $j=1,\dots,m$, is a control input defined by $\mathbf{a}_{j} = \left[ V, \omega_{j} \right]^T $. For an action $\mathbf{a}_{j} \in A$, the expected distance towards the goal position, $\mathbf{x}_{f} = [x_{f}~y_{f}]^{T}$, is defined by
\begin{equation} 
    d(\mathbf{a}_{j}) = \| \hat{\mathbf{x}}_{k+1} - \mathbf{x}_{f} \|
    \label{eq:exp_dist}
\end{equation}
where
\begin{equation}
    \hat{\mathbf{x}}_{k+1} = \begin{bmatrix}
        x_{k} + V\cos(\theta_{k}+\omega_{j}\Delta t)\\
        y_{k} + V\sin(\theta_{k}+\omega_{j}\Delta t)
    \end{bmatrix}
\end{equation}
represents the expected position of the robot when $\mathbf{a}_{j}$ is executed. Then, the cost function to consider the aforementioned trade-off is designed by
\begin{equation}
    J(\mathbf{a}_{j}) = w_{h} \cdot \alpha^{EER(\mathbf{a}_{j})} + w_{d} \cdot d(\mathbf{a}_{j})
    \label{eq:cost_function}
\end{equation}
where $\alpha$ is a constant set by users depending on applications to scale EER values. The weights $w_{h}$ and $w_{d}$ are also chosen by users depending on the importance between the localization uncertainty reduction and travel time efficiency. Then, the presented uncertainty-aware motion planning algorithm computes the action by
\begin{equation}\label{eqn:min_cost}
    \mathbf{u}_{k} = \arg \min_{\mathbf{a}_{j}\in A} J(\mathbf{a}_{j})
\end{equation}

\subsection{Real-time Implementation via PDF Approximation}
\label{ssec:particle_approximation}
% why we do this: arbitrary shape of distribution
In order to enable real-time onboard planning for magnetic-based navigation, the presented work extends the concept of particle filter to approximate probability distributions for information-driven motion planning. There are two main reasons why this approximation is the best suitable approach. First, the probability distribution in robot localization $bel(\mathbf{x}_{k})$ depends on the shape of magnetic field map. Specifically, the belief is updated based on the magnetometer reading and where that reading could be obtained from the map. Thus, the distribution $bel(\mathbf{x}_{k})$ is often highly non-Gaussian and arbitrary.

% why we do this 2: computational complexity
Second, this approximation can be used to approximate values of information-theoretic function in planning stage. Computing an expected information gain is known to be computationally extensive as the expectation is computed by considering all possible measurements for each possible action. Since entropy calculation in the continuous domain becomes highly complex, the robot states, future states, and sensor measurements can all be represented as a set of particles \cite{stachniss_information_2005}. Thus, the presented approximation method not only resolves the issues with non-Gaussian distributions but also enables real-time computation of expected information gain values by limiting the number of particles when computing resources are limited in hardware setups.

% entropy approximation
Based on \cite{boers_particle_2010}, entropy in \eqref{eq:entropy} can be approximated by a particle-based method by
\begin{multline}
     H\left( p\left(\mathbf{x}_k \mid z_k \right)\right) \approx \log \left( \sum_{i=1}^{N}  p \left( z_k \mid \mathbf{x}^{[i]}_{k} \right) w^{[i]}_{k-1} \right) \\ -\sum_{i=1}^{N} \log\Bigg( p(\left(z_k \mid \mathbf{x}_k^{[i]}\right)
     \cdot\left(\sum_{j=1}^{N} p\left(\mathbf{x}_k^{[i]}\mid \mathbf{x}_{k-1}^{[j]}\right) w^{[j]}_{k-1} \right) w_k^{[i]} \Bigg)
     \label{eqn:entropy_current_particle}
\end{multline}
where $N$ represents the number of particles (or hypothesis) used in approximation and $w_{k}^{[i]}$ represents the weight of the $i$th particle at time $k$. When sensor measurement is unavailable, this equation reduces as follows \cite{rui_path_2018}
\begin{multline}
    H\left( bel\left(\mathbf{x}_k \right)\right) \approx \\-\sum_{i=1}^{N} \log\Bigg(\left(\sum_{j=1}^{N} p\left(\mathbf{x}_k^{[i]}\mid \mathbf{x}_{k-1}^{[j]}\right) w^{[j]}_{k-1} \right) w_k^{[i]} \Bigg)
    \label{eqn:entropy_current_particle_no_measurement}
\end{multline}

By applying the similar approach, the entropy in future time step, which needs to be approximated for real-time implementation, can be represented by
\begin{multline} 
    H\left(bel\left(\mathbf{x}_{k+1} \mid \mathbf{x}_k , u_k, z_{k+1} \right) \right) \approx 
    \log \left(\sum_{i=1}^{M} p\left(z_{k+1}\mid \mathbf{x}^{[i]}_{k+1}\right) w^{[i]}_{k} \right) 
    \\ -\sum_{i=1}^{M}\log\Bigg(p\left(z_{k+1} \mid \mathbf{x}_{k+1}^{[i]}\right)
    \cdot\left(\sum^{M}_{j=1} p\left(\mathbf{x}_{k+1} \mid \mathbf{x}_k^{[i]}\right)w^{[j]}_k\right) w^{[i]}_{k+1}\Bigg)
    \label{eqn:entropy_future_particle}
\end{multline}
where $p\left(z_{k+1}\mid \mathbf{x}^{[i]}_{k+1}\right)$ can be calculated for the future pose using the measurement model \eqref{eq:sensor_model}. Then, EER in the planning algorithm can be approximated and calculated by
\begin{multline} 
    EER\left(u_k\right) = H\left( bel\left(\mathbf{x}_k \right)\right) \\ - \sum^{M}_{j=1}  p\left(z_{k+1} \mid \mathbf{x}^{[j]}_{k+1} \right) \cdot   H\left(bel\left(\mathbf{x}^{[j]}_{k+1} \mid \mathbf{x}_k , \mathbf{u}_k, z_{k+1} \right) \right)
    \label{eqn:EER_particle}
\end{multline}
where $M$ is the number of EER hypotheses used in the approximation. Equation \eqref{eq:sensor_model} is the representative future measurement found by querying the map at the $i$th particle.

\section{EXPERIMENT SETUP}
\label{sec:experiment_setup}
\begin{figure}[t]
    \centering
    \subfloat[\label{fig:synthetic_map}]{%
        \includegraphics[width=0.48\linewidth]{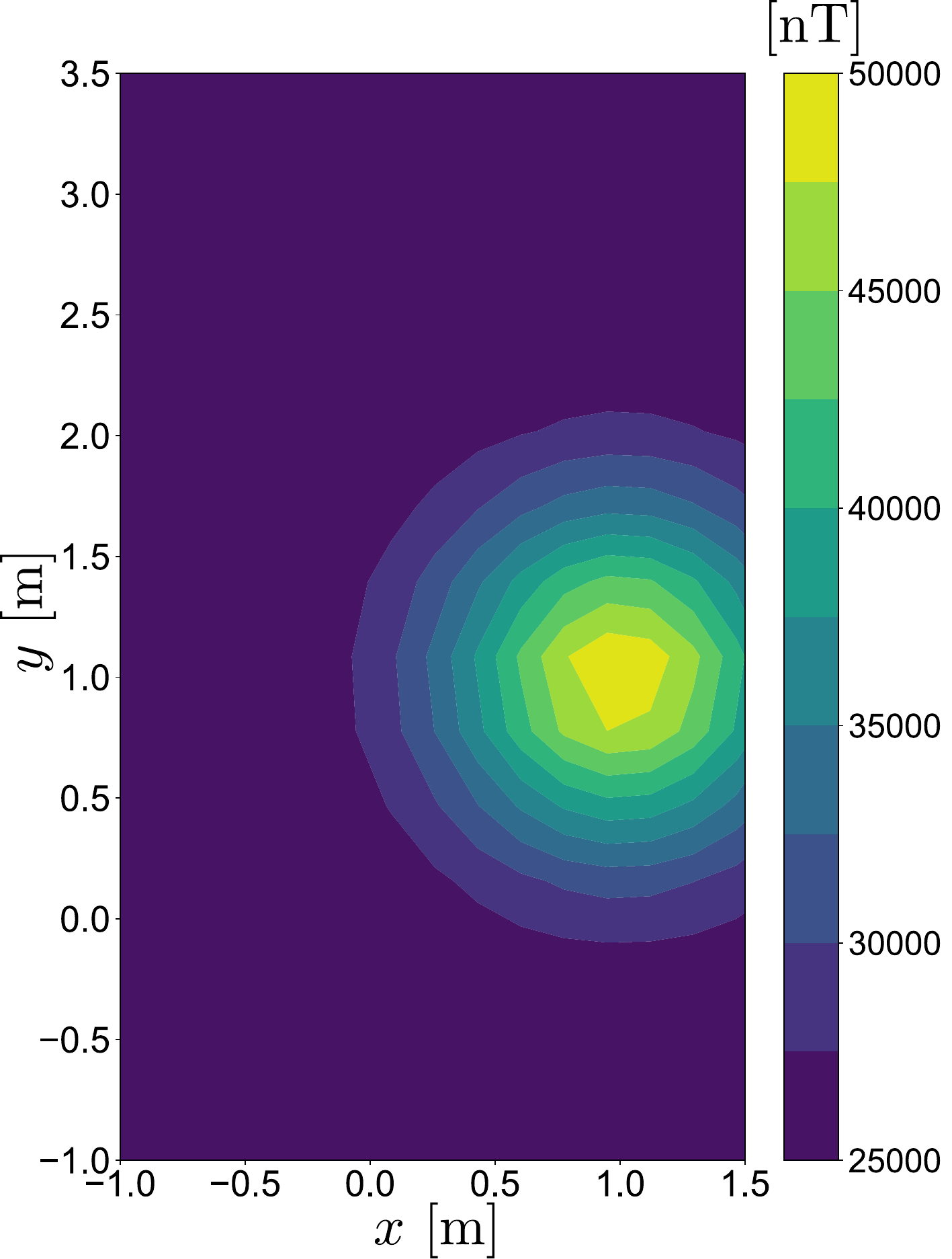}}
    \hfill
    \subfloat[\label{fig:exp_map}]{%
        \includegraphics[width=0.48\linewidth]{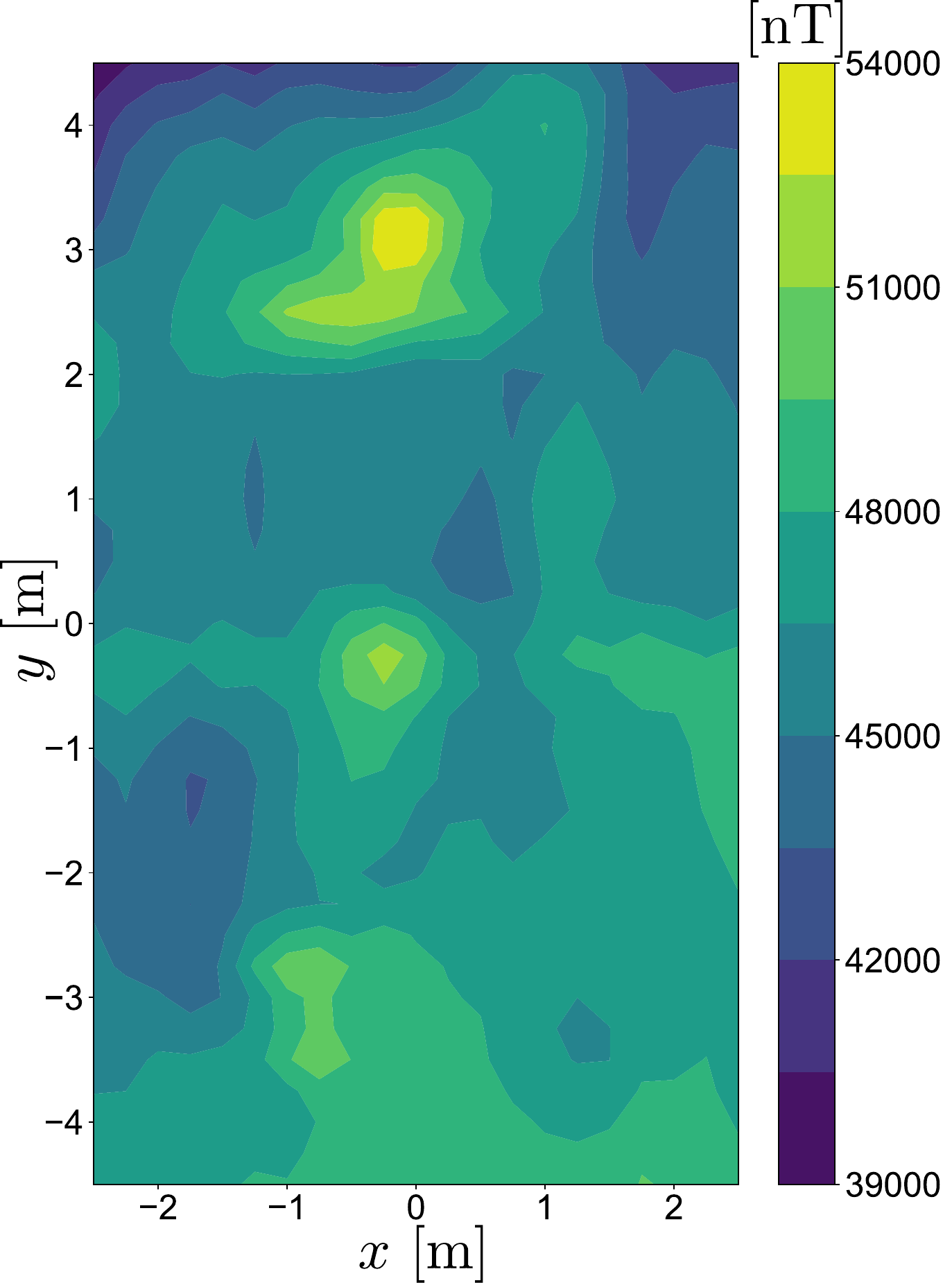}}
    \caption{The magnetic field maps used for (a) simulation and (b) hardware demonstration}
    \label{fig:maps}
\end{figure}

Both simulation and hardware experiments have been performed to demonstrate the proposed real-time motion planning algorithm for magnetic-based navigation. For the simulation experiments, a simple synthetic magnetic field map (Fig. \ref{fig:synthetic_map}) is created by adding a multivariate Gaussian with a mean of $25,000\mathrm{nT}$ and standard deviation of $1000 \mathrm{nT}$ into a uniform value of $25,000 \mathrm{nT}$ to create a single peak. For hardware demonstration, the magnetic field map of an indoor experimental space (Fig. \ref{fig:exp_map}) is obtained by collecting total-field magnetometer readings with exact robot locations from a motion capture system, while the robot is stationary at each measurement point to minimize the magnetic interference from actuation motors. These map acquisition and experiments are performed using a TurtleBot equipped with Twinleaf Micro-SAM and VMR as the scalar and vector magnetometers for Tolles-Lawson calibration.

The same objective function and user-chosen parameters have been used in both simulation and hardware experiments. The objective function in \eqref{eq:cost_function} is set by $\alpha = 0.9$ and $w_{d} = \frac{1}{500}$ considering the scale of $EER$ and distance values. In the planning algorithm, the action set $A$ consists of six actions $\mathbf{a}_{j}=[V,\omega_{j}]^{T}$, for $j=1,\dots,6$, with the linear velocity $V=0.2~[\text{m/s}]$ and angular velocity $\omega_{j} \in [-25,-15,-5,5,15,25]~[\text{deg/s}]$. For the measurement model \eqref{eq:sensor_model}, the scalar magnetometer variance is set to be $(150 \mathrm{nT})^{2}$. For the motion model \eqref{eq:motion_model}, the covariance is set by $diag\{0.01~[\text{m}], 0.01~[\text{m}], 0.15~[\text{deg}]\}^{2}$. For the particle filter implementation in localization, $N=250$ particles are used with resampling threshold of $\frac{N}{2}$. For EER estimation, $M=30$ particles are used to predict the entropy reduction in 10 time steps ahead, where the update frequency is set by $10[\text{ Hz}]$.

\section{RESULTS AND DISCUSSIONS}
\label{sec:results_discussions}

\subsection{Simulation Experiment Results}
\label{ssec:sim_results}
\begin{figure}[tb]
    \centering
    \includegraphics[width=0.65\columnwidth]{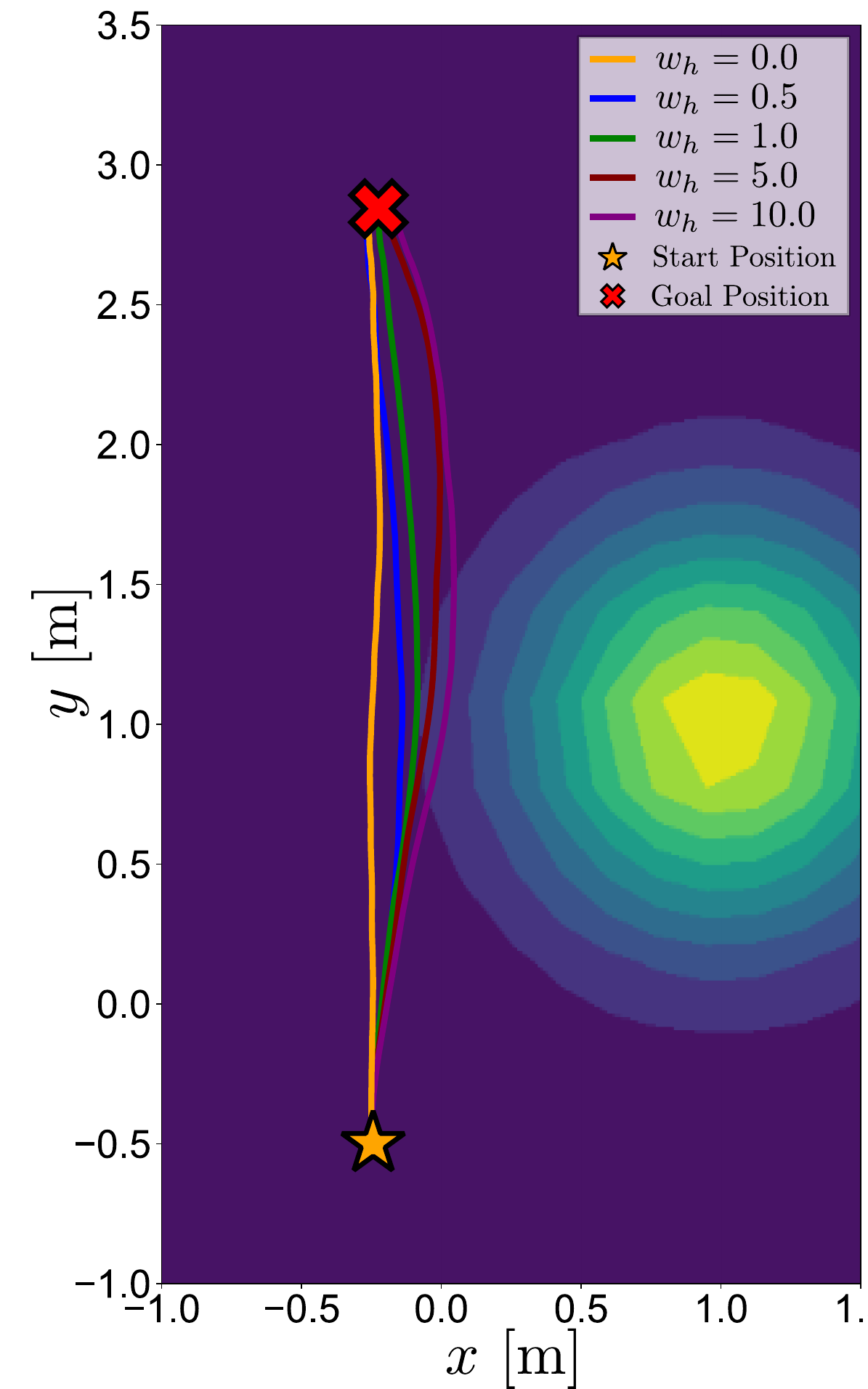}
    \caption{Comparison of the robot trajectories executed by the presented algorithm with weights $w_{h}=0$, $0.5$, $1$, $5$ and $10$
    % A composite image of the ground truth paths during simulation. The weights in this image are swept from $W_{EER}=0.0$ to $10.0$. As the weights tend to be higher the further the path navigates through the gradient of the magnetic anomaly.
    }
    % fix the legend and manage the font size in the axis labels and ticks
    \label{fig:synthetic_composite}
\end{figure}

Simulation experiments are performed to demonstrate the characteristics of the proposed motion planning algorithm. As described in Section \ref{sec:experiment_setup}, an artificial magnetic field map consisting of a symmetric Gaussian peak is used. A range of weights in the objective function, from $w_{h} = 0$ to $10$, are used to generate trajectories to show the impact of EER-based term in the objective function, with $w_{d}$ fixed. The generated trajectories are plotted in Fig. \ref{fig:synthetic_composite} for comparison. The results show that, as the planning algorithm considers uncertainty measure more with a higher $w_{h}$ value, the algorithm guides the robot further into the magnetic field's gradient on the right side of the map. When $w_{h}=0$, which means that the localization uncertainty is not considered, the robot trajectory is almost a straight line.

% \begin{figure*}[t]
%     \centering
%     \subfloat[\label{fig:synthetic_composite}]{%
%         \includegraphics[width=0.6\columnwidth]{figures/FigSimResultsTrajectoriesComposite.pdf}}
%     \hfill
%     \subfloat[\label{fig:sim_entropy_cov_comparison}]{%
%         \includegraphics[width=1.3\columnwidth]{figures/FigSimResultsTrajectoriesEntropyCov.pdf}}
%     \caption{(a) A composite image of the ground truth paths during simulation. The weights in this image are swept from $W_{EER}=0.0$ to $10.0$. As the weights tend to be higher the further the path navigates through the gradient of the magnetic anomaly, (b) \js{Aditya - update the figures.}\ap{Updated}}
%     \label{fig:sim_results} 
% \end{figure*}

\begin{figure}[t]
    \centering
    \includegraphics[width=0.99\linewidth]{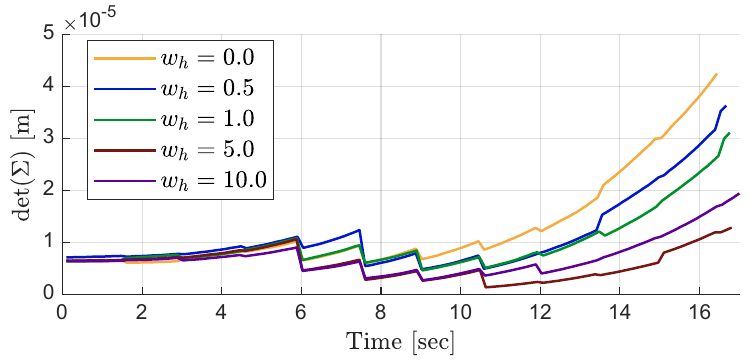}
    \caption{Comparison of the determinant of covariance matrix over time, with weights $w_{h}=0$, $0.5$, $1$, $5$ and $10$}
    \label{fig:synthetic_composite_uncertainty}
\end{figure}

The detailed results of localization uncertainty is shown in Fig. \ref{fig:synthetic_composite_uncertainty} in terms of the determinant of covariance matrix over travel time. It is shown that, when the localization uncertainty is weighted higher in the objective function (higher $w_{h}$), the covariance of the estimation reduces significantly more, leading to a lower covariance over time. In addition, these results validate that visiting the gradient area in the map gives higher localization uncertainty reduction and thus lower localization uncertainty represented in covariance.

\begin{figure}[t]
    \centering
    \includegraphics[width=0.6\linewidth,trim={0 0 25cm 0},clip]{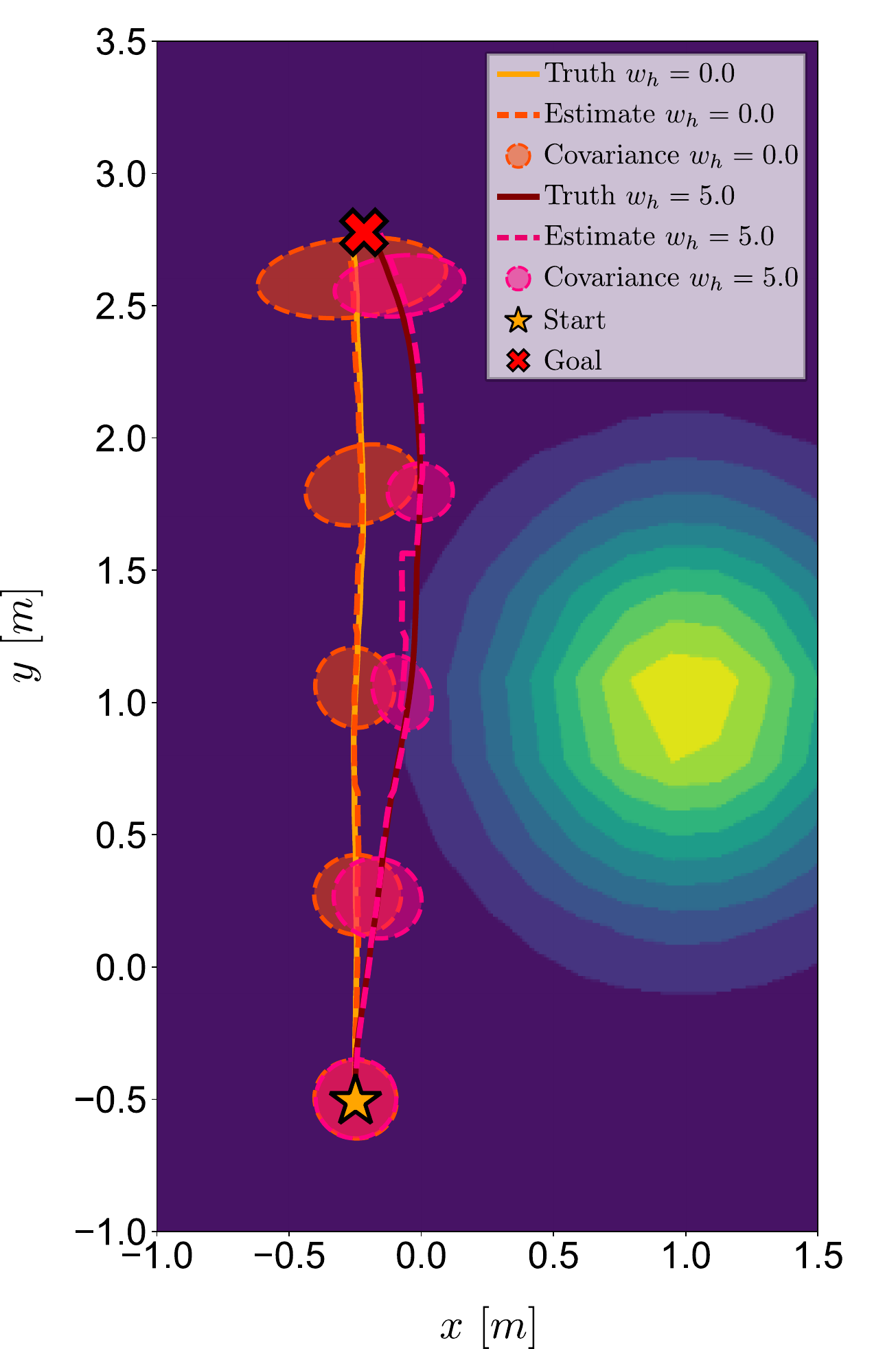}
    \caption{The comparison between two experimental runs with $w_{h}=0$ and $5$ by the ground truth trajectory (solid line), estimated trajectory (dashed line), and covariance (oval)}
    \label{fig:synthetic_comparison_trajectory}
\end{figure}

In order to directly compare two trajectories and corresponding localization uncertainties over time, Fig. \ref{fig:synthetic_comparison_trajectory} shows the two trajectories with covariance on the map when $w_{h}=0$ and $w_{h}=5$. It is clearly shown that the trajectory generated with $w_{h}=0$ results in a bigger covariance due to the uniform magnetic values obtained from the area that the straight trajectory passes by. On the contrary, when $w_{h}=5$, the covariance starts to increase at the beginning, however, as the robot is guided through the edge of the gradient, the covariance significantly improves.

\subsection{Hardware Experiment Results}
\label{ssec:exp_results}
\begin{figure}
    \centering
    \subfloat[\label{fig:exp_trajectory_wh_0}]{%
        \includegraphics[width=0.48\columnwidth]{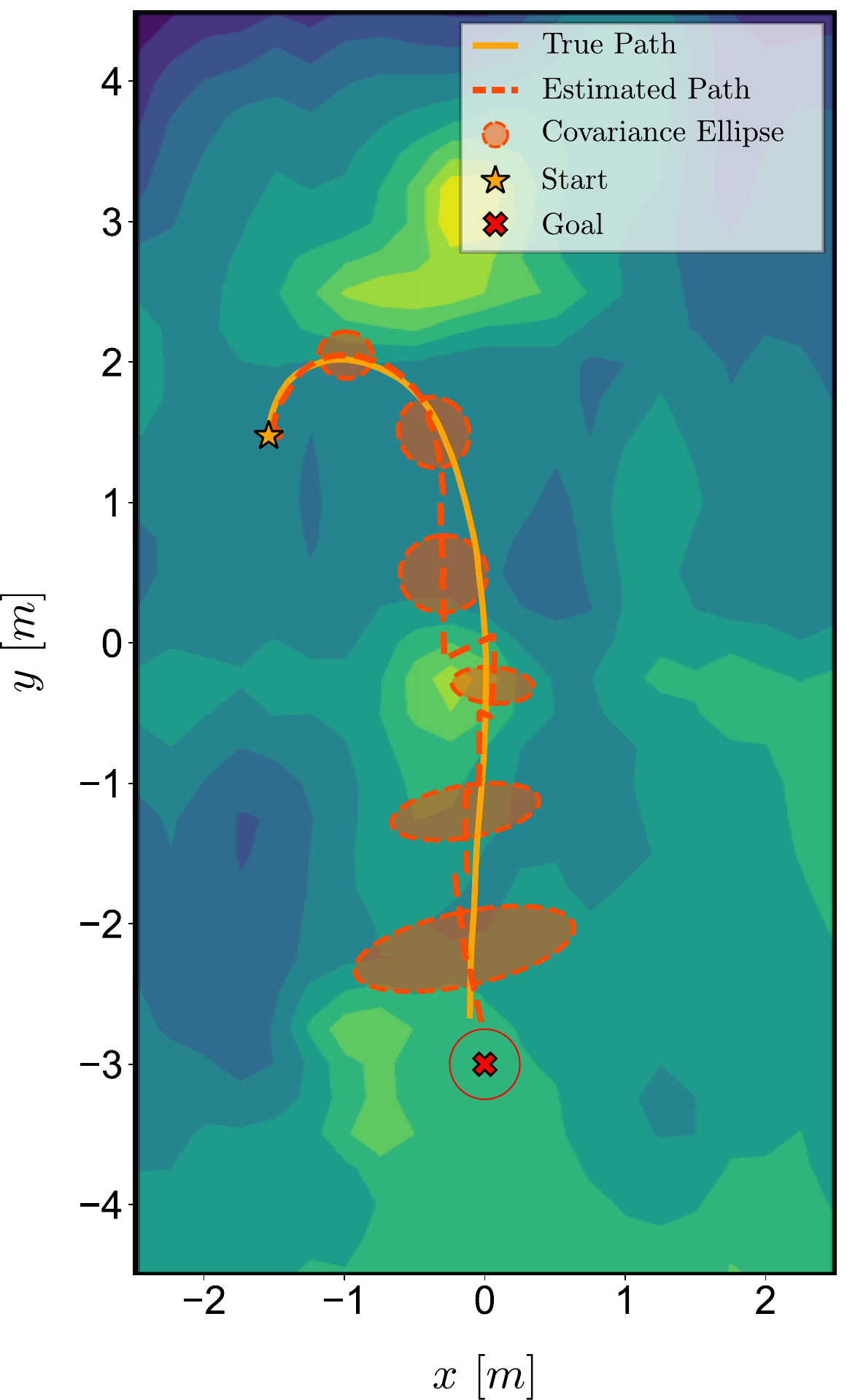}}
    \subfloat[\label{fig:exp_trajectory_wh_05}]{%
        \includegraphics[width=0.48\columnwidth]{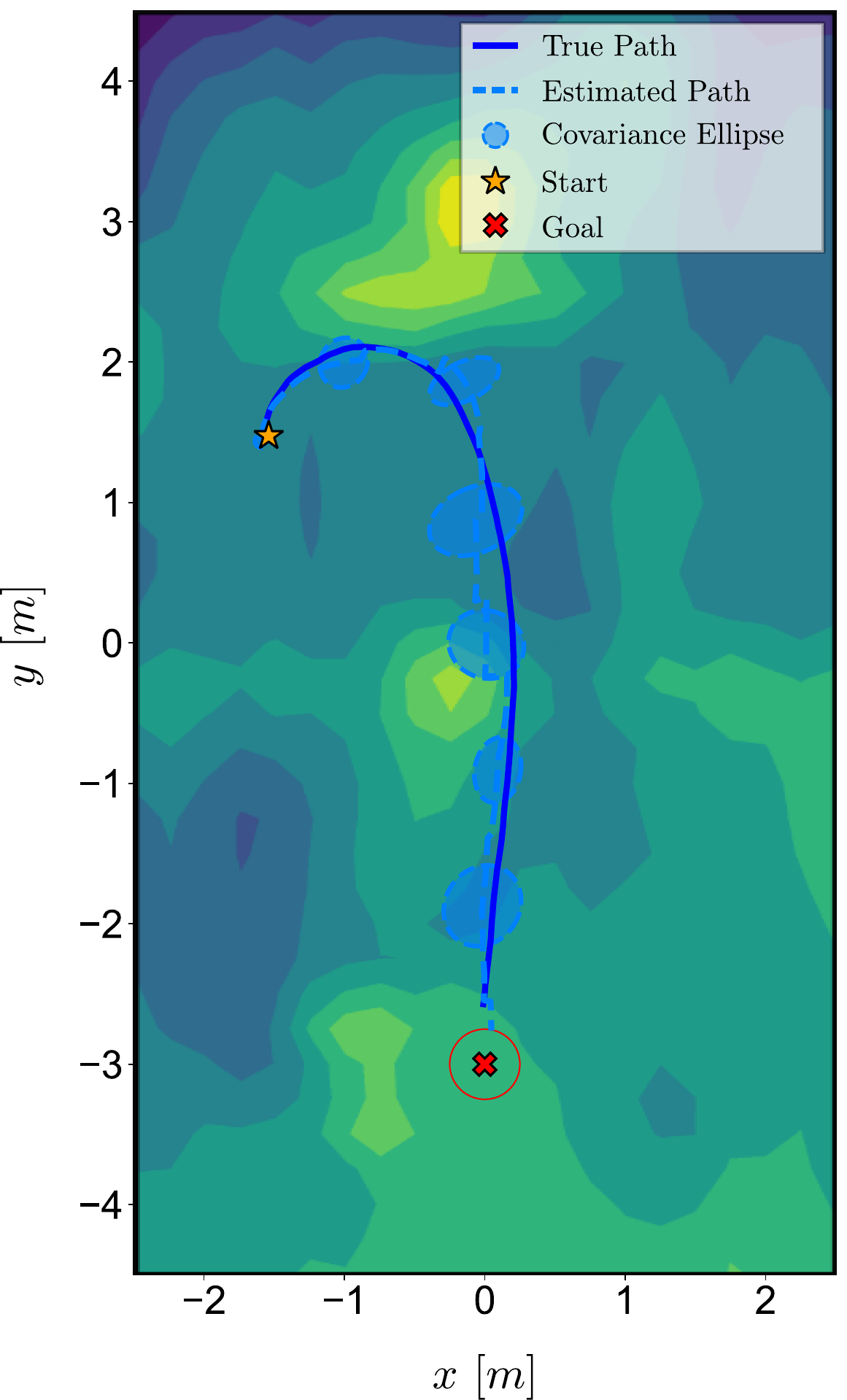}}
    \\
    \subfloat[\label{fig:exp_trajectory_wh_1}]{%
        \includegraphics[width=0.48\columnwidth]{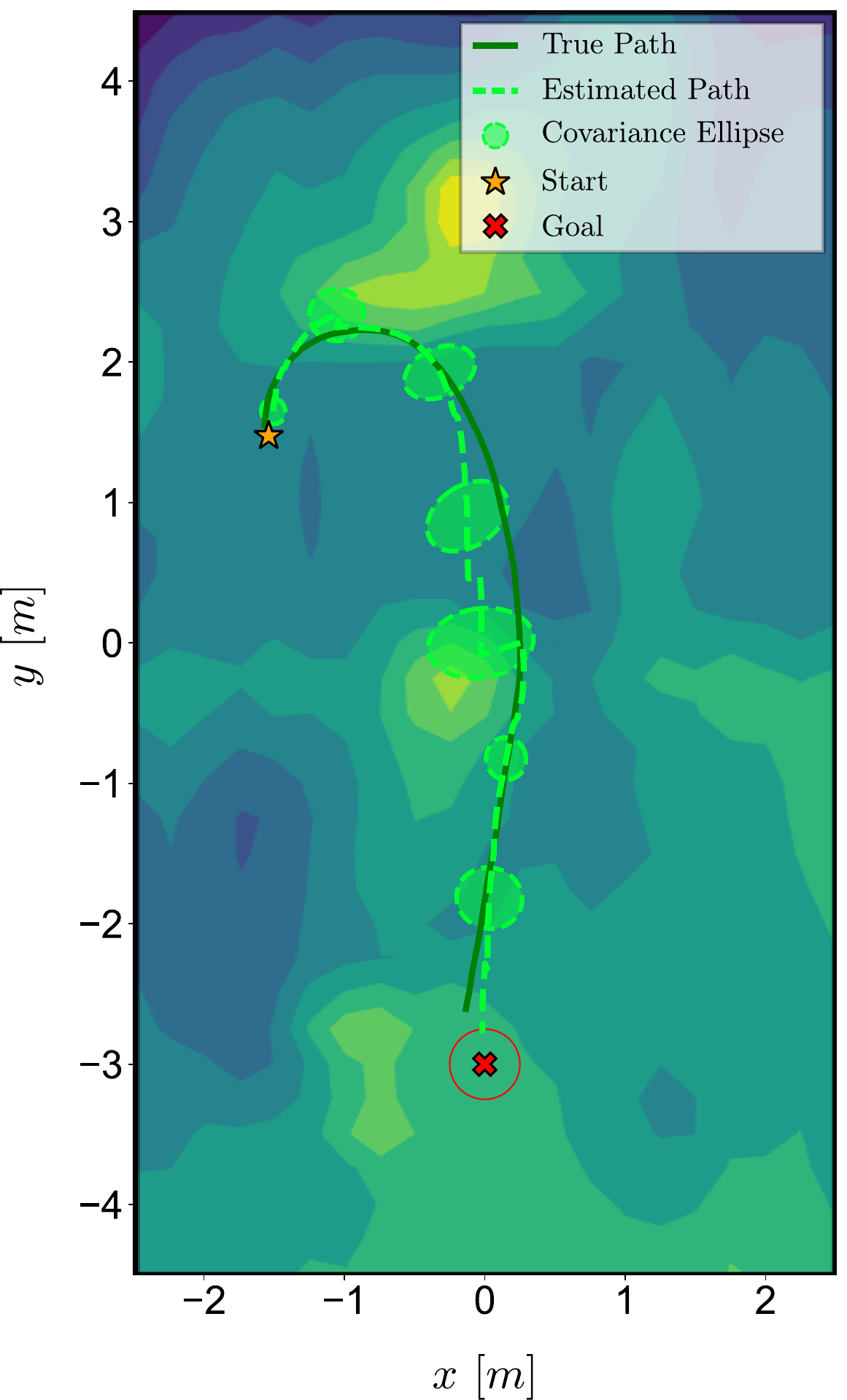}}
    \subfloat[\label{fig:exp_trajectory_wh_5}]{%
        \includegraphics[width=0.48\columnwidth]{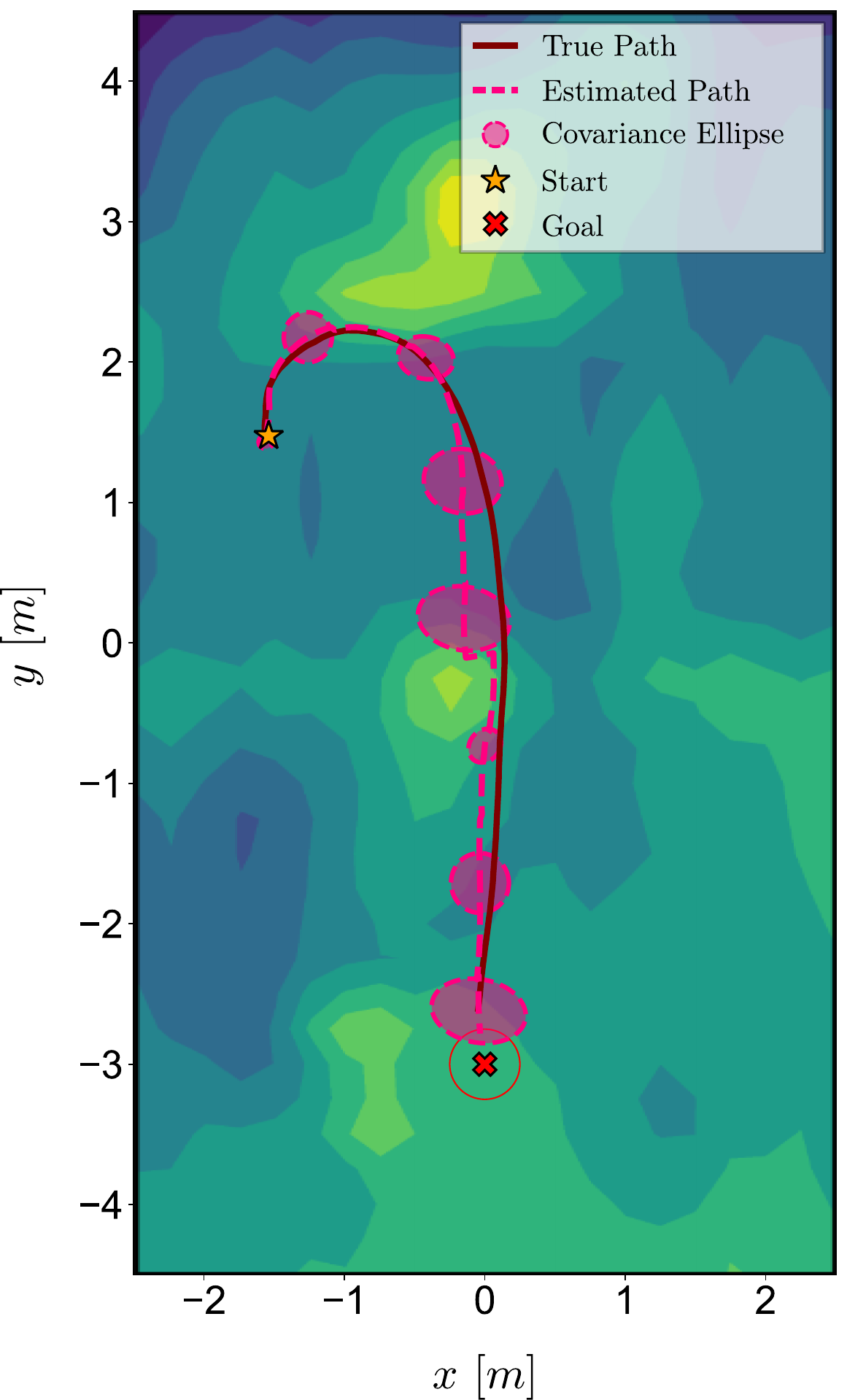}}
    \caption{The executed trajectory and localization covariance when (a) $w_h=0$, (b) $w_h=0.5$, (c) $w_h=1$, and (d) $w_h=5$}
    \label{fig:exp_trajectories} 
\end{figure}

The presented uncertainty-aware planning approach is demonstrated with a hardware experiment in an indoor lab using a ground robotic platform described in Section \ref{sec:experiment_setup}. The hardware experiments have validated that the presented guidance method runs in real-time with an update frequency of 10Hz with the basic robotic platform. Similarly to the simulation experiments, the presented method has been tested with a range of weights in the objective function, from $w_{h} = 0.0$ to $5.0$, in order to examine the behavior of considering localization uncertainty. The results are shown in Fig. \ref{fig:exp_trajectories}, where the trajectory and corresponding covariance from localization (i.e., state estimation using particle filter) are plotted for each test $w_{h}$ value.

% \begin{figure*}
%     \centering
%     \subfloat[\label{fig:exp_trajectory_wh_0}]{%
%         \includegraphics[width=0.48\columnwidth]{figures/FigExpResultsTrajectoryWh0.pdf}}
%     \hspace{8pt}\subfloat[\label{fig:exp_trajectory_wh_05}]{%
%         \includegraphics[width=0.48\columnwidth]{figures/FigExpResultsTrajectoryWh05.pdf}}
%     \hspace{8pt}\subfloat[\label{fig:exp_trajectory_wh_1}]{%
%         \includegraphics[width=0.48\columnwidth]{figures/FigExpResultsTrajectoryWh1.pdf}}
%     \hspace{8pt}\subfloat[\label{fig:exp_trajectory_wh_5}]{%
%         \includegraphics[width=0.48\columnwidth]{figures/FigExpResultsTrajectoryWh5.pdf}}
%     \caption{The executed trajectory and localization covariance when (a) $w_h=0$, (b) $w_h=0.5$, (c) $w_h=1$, and (d) $w_h=5$.}
%     \label{fig:exp_trajectories} 
% \end{figure*}

As illustrated in Fig. \ref{fig:exp_trajectories}, the planning algorithm directs the robot towards the peak near the start position as the weight $w_{h}$ increases. Initially, the robot is oriented along the y-axis, necessitating a turn to head towards the goal position. When  $w_{h}=0$, as shown in Fig. \ref{fig:exp_trajectory_wh_0}, the ground robot executes a sharp turn to directly approach the goal position. However, with increasing  $w_{h}$, the robot traverses areas with higher magnetic field gradients to achieve greater information gain, thereby reducing localization uncertainty. This traversal is beneficial because the magnetometer readings in these regions are distinct from those in nearby areas. Notably, as $w_{h}$  increases, the robot achieves a more reasonable localization uncertainty. Specifically, the covariance increases with larger estimation errors and decreases with smaller estimation errors, reflecting an adaptive uncertainty estimation based on the robot's information gain.

\begin{figure} 
    \centering
    \subfloat[\label{fig:exp_entropy}]{%
        \includegraphics[width=0.98\linewidth]{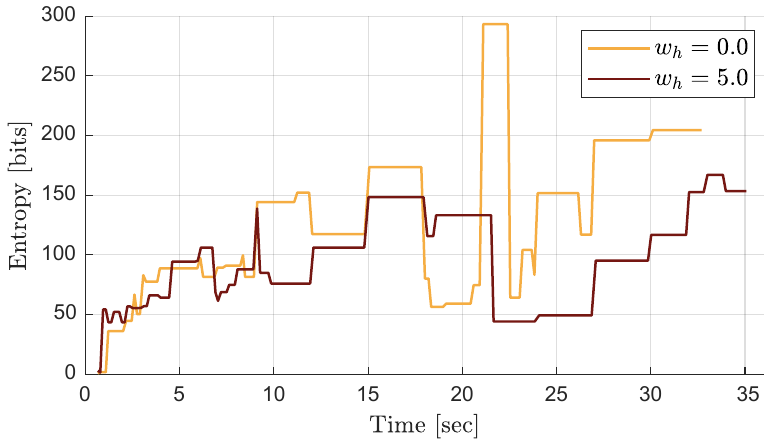}}
        \\
    \subfloat[\label{fig:exp_cov}]{%
        \includegraphics[width=0.98\linewidth]{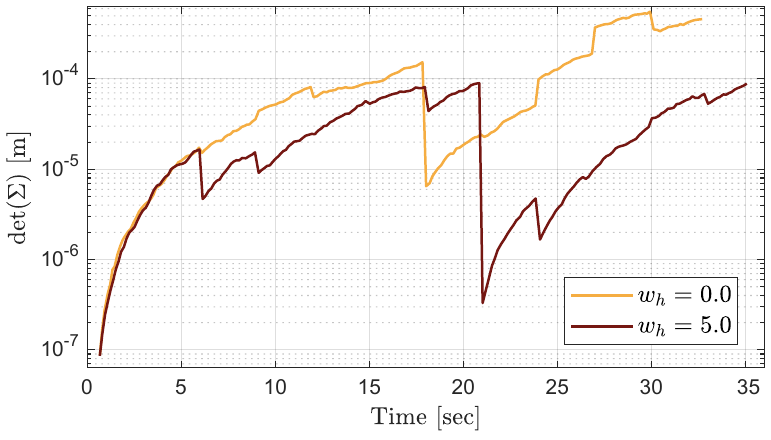}}
    \caption{Comparison of the localization uncertainty from the two trajectories executed with $w_{h}=0$ and $w_{h}=5$ in terms of (a) entropy and (b) determinant of covariance matrix}
    \label{fig:hardware_exp_entropy_cov} 
\end{figure}

The detailed results on localization uncertainty for two trajectories with $w_{h}=0$ and $w_{h}=5$ are shown in Fig. \ref{fig:hardware_exp_entropy_cov}. In Fig. \ref{fig:exp_entropy}, the entropy over path for hardware experiments is depicted. Specifically, an outlier rejection methodology using 1.5 times the interquartile range (IQR) has been applied over this data to reduce spikes caused by resampling in particle filter as those peaks are not considered in navigation. In Fig. \ref{fig:exp_cov}, the corresponding determinant of covariance matrix is shown from the hardware experiments. The results show that the localization uncertainty reduces significantly when the robot passes by the area with magnetic field gradient. 

While the hardware experiments demonstrate the real-time implementation of the presented planning algorithm for MagNav, there are several challenges coming from the entropy approximation. The precision of the state entropy and by extension, the information gain is inherently dependent on the the number of particles, distribution of weights, and the resampling of the particles and weights. The number of particles directly affects the precision of the entropy, as discussed in \cite{boers_particle_2010}. As the number of particles is increased, entropy approximation converges to the true entropy value calculated from the distribution. The distribution of the weights and resampling step affect this calculation. Through observation, if the weights are uniformly distributed, the entropy rises unboundedly, giving a poor approximation. A similar problem is faced during particle resampling, as the weights are re-initialized as uniform, which causes the spikes in entropy seen in Fig. \ref{fig:exp_entropy} of Fig. \ref{fig:hardware_exp_entropy_cov}.

\section{CONCLUSIONS}
In this work, a novel real-time uncertainty-aware motion planning algorithm is proposed and demonstrated such that the robot's localization uncertainty reduces in GPS-denied navigation scenarios by utilizing a known magnetic field map and onboard magnetometer sensors. Utilizing the given map in the sensor model, the presented planning algorithm guides the robot to obtain useful information from the magnetometers to reduce localization uncertainty while moving towards the goal position. The real-time implementation is enabled by the approximation of the expected entropy reduction method that extends the particle filter concept for planning. This novel planning approach is evaluated and demonstrated via both simulations and hardware experiments. Both the simulation and hardware results demonstrate the effectiveness of the presented algorithm to guide the robot towards the area with a high information for localization, leading towards significant reduction in localization uncertainty. As a future work, the authors will include a global path planner to leverage the information-rich magnetic maps and address the local nature of the current receding horizon guidance law. This future work will address current limitation in finding only locally optimal paths and make the presented method more suitable for long duration operation.

\addtolength{\textheight}{-12cm}   % This command serves to balance the column lengths

%%%%%%%%%%%%%%%%%%%%%%%%%%%%%%%%%%%%%%%%%%%%%%%%%%%%%%%%%%%%%%%%%%%%%%%%%%%%%%%%

%%%%%%%%%%%%%%%%%%%%%%%%%%%%%%%%%%%%%%%%%%%%%%%%%%%%%%%%%%%%%%%%%%%%%%%%%%%%%%%%



% \section*{ACKNOWLEDGMENT}

% The preferred spelling of the word ÒacknowledgmentÓ in America is without an ÒeÓ after the ÒgÓ. Avoid the stilted expression, ÒOne of us (R. B. G.) thanks . . .Ó  Instead, try ÒR. B. G. thanksÓ. Put sponsor acknowledgments in the unnumbered footnote on the first page.

%%%%%%%%%%%%%%%%%%%%%%%%%%%%%%%%%%%%%%%%%%%%%%%%%%%%%%%%%%%%%%%%%%%%%%%%%%%%%%%%

\bibliography{magnav}
\bibliographystyle{IEEEtran}

\end{document}